\documentclass{article}
\usepackage[final]{corl_2026} %

\usepackage{moreverb,url}
\usepackage{xspace}
\usepackage{xcolor}

\usepackage{amssymb}
\usepackage{amsmath}
\usepackage{amsbsy}
\usepackage{bbm}
\usepackage{dsfont}

\usepackage{graphicx}
\usepackage{wrapfig}
\usepackage{booktabs} %
\usepackage{subcaption}

\usepackage{subfiles}
\usepackage{csquotes}
\usepackage{makecell} %

\usepackage{array}

\usepackage{graphicx}

\usepackage{multirow}

\usepackage{float}
\usepackage{amsmath,amssymb,amsfonts}
\usepackage{algorithmic}
\usepackage{graphicx}
\usepackage{textcomp}
\usepackage{xcolor}
\usepackage{booktabs}

\definecolor{wine}{RGB}{204, 0, 102}
\definecolor{magenta_wine}{RGB}{158, 44, 143}
\definecolor{dusty_wine}{RGB}{143, 59, 101}
\definecolor{ocean}{RGB}{12, 119, 157}
\definecolor{light_ocean}{RGB}{18, 178, 235}
\definecolor{dark_ocean}{RGB}{10, 89, 148}
\definecolor{grey}{RGB}{170, 170, 170}
\definecolor{light-grey}{RGB}{220, 220, 220}
\definecolor{dark-grey}{rgb}{0.2, 0.2, 0.2} 
\definecolor{med-grey}{rgb}{0.3, 0.3, 0.3} 
\definecolor{grape}{RGB}{112,48,160}
\definecolor{aqua}{RGB}{52,172,139}
\definecolor{dark_aqua}{RGB}{35,115,93}
\definecolor{dark_orange}{RGB}{216,92,0}
\definecolor{golden_orange}{RGB}{234, 136, 0}
\definecolor{vibrant_orange}{RGB}{250, 160, 26}
\definecolor{vibrant_blue}{RGB}{14, 120, 255}
\definecolor{vibrant_pink}{RGB}{255, 0, 104}
\definecolor{dark_red}{RGB}{122, 0, 0}
\definecolor{dark_green}{RGB}{0, 92, 34}
\definecolor{dusty_blue}{RGB}{77, 91, 128}
\definecolor{dark_brown}{RGB}{125, 54, 36}

\newcommand{\para}[1]{\medskip\noindent\textbf{#1. }} 
 
\newcommand{\paranopunc}[1]{\medskip\noindent\textbf{#1 }} %

\newcounter{qnum}
\newcommand{\wmrgb}{$\textbf{WM}_\textbf{RGB}$}
\newcommand{\wmmm}{$\textbf{WM}_\textbf{MM}$}
\newcommand{\wmmasked}{$\textbf{WM}_\textbf{RGB-MM}$}

\newcommand{\latent}{z}
\newcommand{\latentSpace}{\mathcal{Z}}

\newcommand{\dyn}{f}

\newcommand{\ellparam}{\mu}

\newcommand{\obs}{o}
\newcommand{\obsSpace}{\mathcal{O}}
\newcommand{\enc}{\mathcal{E}}

\newcommand{\action}{a}
\newcommand{\actionSpace}{\mathcal{A}}

\newcommand{\policyTask}{\pi^{\text{nom}}}

\usepackage{fontawesome5}
\newcommand{\shield}{\text{\tiny{\faShield*}}}

\newcommand{\marginfunc}{\ell}

\newcommand{\wax}{{\textcolor{ocean}{\textbf{Wax Melting}}}\xspace} %
\newcommand{\rice}{{\textcolor{golden_orange}{\textbf{Rice Pouring}}}\xspace} 
\usepackage{annotate-equations}

\author{
  Matthew Kim$^{*1}$, Kensuke Nakamura$^{*2}$, Andrea Bajcsy$^2$ \\
  $^{1}$UC San Diego \;\;\; $^{2}$Carnegie Mellon University \;\;\;  $^*$Equal contribution\\{\tt\footnotesize mak009@ucsd.edu, \{kensuken, abajcsy\}@andrew.cmu.edu}
}

\begin{document}

\title{\LARGE \bf 
How Well Do Latent World Models Understand Partially Observable Safety Constraints?

}

\maketitle

\begin{abstract}
Latent world models are a promising approach for learning state representations and dynamics directly from high-dimensional observations, enabling robot control in hard-to-model settings. 
However, control performance ultimately depends on the latent representation encoding the required information for the task.
In this work, we study latent-space safe control problems and show how partial observability can induce control failures when safety-relevant information is \textit{not} preserved in the latent state.
Specifically, we identify two world model failure modes: \textcolor{black}{\textit{estimation gaps}}, where current observations do not reveal safety-critical quantities (e.g., temperature in a cooking task), and \textcolor{black}{\textit{prediction gaps}}, where failures are observable once they occur but cannot be reliably anticipated from available observations.
We introduce two diagnostics for these gaps: a mutual-information-based measure of safety observability and a rollout-based measure of future safety predictability. 
Finally, we present mitigation strategies for each failure mode: privileged multimodal supervision for estimation gaps and conformal risk calibration for prediction gaps. 
Across two hardware case studies---using unimodal RGB world models and multimodal RGB+Tactile and RGB+Thermal variants---we show that these mitigation strategies improve the safety of a Franka Research 3 manipulator on challenging cooking tasks under partial observability, albeit with increased conservativeness.
More broadly, our work raises the question of when world model state representations are sufficient for reliable robot control.
\textbf{Project Webpage:} \url{https://cmu-intentlab.github.io/multisafe}

\end{abstract}

\section{Introduction}

Latent world models (WMs)~\cite{ha2018world} offer a promising alternative to hand-specified dynamics models and support downstream planning~\cite{hafner2019learning, zhou2024dino}, policy evaluation~\cite{veorobotics2025}, and model-based reinforcement learning \cite{Hafner2020Dream, hansen2024tdmpc2} in scenarios where traditional simulators are difficult to design. 
Given a dataset of high-dimensional observations (e.g., RGB images) and low-level robot actions (e.g., joint velocities), latent WMs learn \textit{(i)} an encoder to estimate a lower-dimensional latent state space and \textit{(ii)} a transition function to evolve the latent state conditioned on robot actions and the current latent state.

Despite promising early results, it remains unclear what limitations arise in latent world models when the training data provides only partial observations of the environment.
In many classical control settings, the state variables (e.g., position, velocities) are specified by a designer. Partial observability then leads to \textit{known unknowns}: while the system is unable to directly observe the state variables, it can maintain uncertainty over the well-defined hypothesis space using state estimators such as Kalman filters \cite{kalman1960}. 
In latent world models, however, the state representation itself is inferred from data collected under partial observability. 
This introduces a different failure mode: \textit{unknown unknowns}, where relevant state quantities may be missing from observations, and therefore %
unrepresented in the learned latent space entirely. 
Thus, the challenge is not only to estimate the hidden state but also to ensure that the learned state space captures the variables necessary for control.

\begin{figure}[t!]
    \centering
\includegraphics[width=0.98\linewidth]{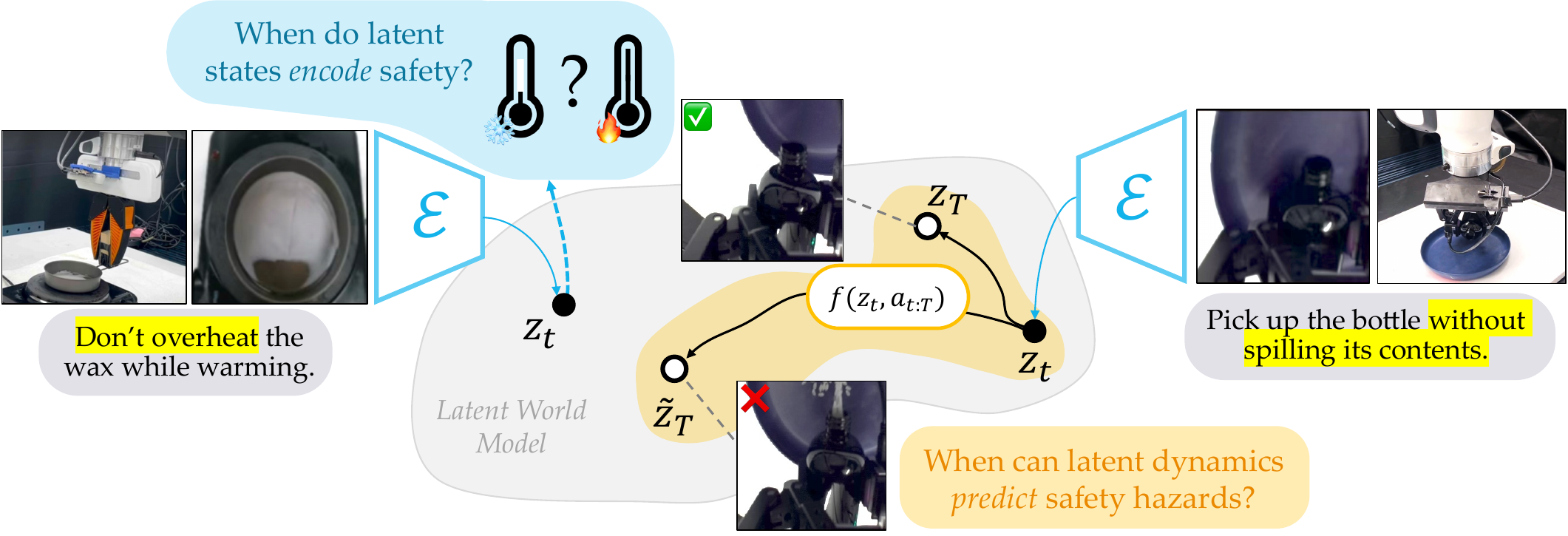}
     \captionof{figure}{\textbf{How well do latent world models (WMs) understand safety constraints?}
     We find that WMs can suffer from \textit{estimation gaps} (left) where the latent state fails to encode safety-relevant states that are hard to observe (e.g., temperature from RGB images) and \textit{prediction gaps} (right) where failures are identifiable from the latent state once they happen (e.g., a spill) but cannot be reliably predicted with current observations because of unobservables (e.g., if the bottle is filled). 
     }
     \vspace{-1em}
     \label{fig:frontfig}
\end{figure}

In this work, we specifically study \textit{latent-space safe control}, a subset of general latent-space control problems where robots must satisfy constraints on latent states throughout time~\cite{wilcox2022ls3, nakamura2025generalizing}.
For example, consider the cooking robot in Figure~\ref{fig:frontfig}. It must heat up, but never overheat, the contents of a pan; it must also pick up, but never spill, the contents of an opaque bottle.
Constraints are represented as classifiers over the latent state (e.g., to detect overheating or a spill), and are %
avoided during safe control optimization via the latent world model predictions.

Our central question is: \textbf{How can we quantify if safety constraints are observable in latent world models, and what mitigation strategies are available if they are not?}
We identify two world model failure modes:
\textcolor{ocean}{\textit{estimation gaps}} %
where observations are insufficient to determine if the system is currently violating a safety constraint, and \textcolor{golden_orange}{\textit{prediction gaps}}, where failures are observable but not reliably predictable from available observations. 
Returning to our cooking example, suppose the robot’s world model and policy are trained using only RGB observations.
When heating a pan, there may be an \textit{estimation gap} because the robot cannot reliably infer the pan's temperature %
from RGB observations alone.
The same robot may suffer from a \textit{prediction gap} when trying to enforce spilling constraints: although it can detect a spill once a bottle's contents have poured out, it cannot reliably predict spills without knowing whether the bottle is filled or empty.

We \textbf{demonstrate}, \textbf{diagnose}, and \textbf{mitigate} these failure modes in hardware experiments on two challenging cooking-inspired tasks (Figure~\ref{fig:frontfig}), using state-of-the-art latent world models~\cite{zhou2024dino, hafner2023dreamerv3}. %
We \textbf{demonstrate} that estimation and prediction gaps both %
degrade latent safety controllers: estimation gaps prevent learned policies from steering away from danger because they poorly capture safety-relevant quantities, while prediction gaps impair future safety anticipation and cause policies to intervene at the wrong time.
We \textbf{diagnose} estimation gaps with a mutual information-based metric that quantifies how much an observation reveals safety violations, and prediction gaps with a rollout-based classification metric. %
We \textbf{mitigate} estimation gaps with a world model intervention that uses multimodal data during training to %
convert \textit{unknown unknowns} into \textit{known unknowns}, and address prediction gaps at the controller level by calibrating the policy's conservativeness based on the world model's failure prediction accuracy.
Our results show that even when only one modality (e.g., RGB) is available at runtime, mitigation strategies can be designed to allow latent safe controllers to %
successfully prevent failures in the face of estimation and prediction observability gaps. 

\vspace{-0.5em}

\para{Statement of Contributions} This paper challenges the assumption of constraint observability in latent world models. First, we identify estimation and prediction gaps as two failure modes that lead latent-space controllers to unsafe behavior. Second, we propose offline diagnostics based on mutual information and classification accuracy to detect these observability gaps from world-model rollouts. Third, we conduct controlled hardware experiments with unimodal and multimodal (RGB+IR and RGB+Tactile) world models for heating and pouring robotic manipulation tasks. Finally, we demonstrate mitigation strategies that improve safety even under RGB-only deployment.

\section{Related Works}

\para{State Variables and State Estimation in Robotics} 
The \textit{state} of a robot system is the collection of quantities necessary to predict the evolution of the system or its reward in the future \cite{cassandras2008introduction}. These state variables are traditionally hand-designed by an expert for a given task, but are rarely known with certainty during deployment. \textit{State estimation} is a subject dealing with the inference of the system's state given a sequence of sensor measurements and a system model~\cite{barfoot}. The Bayes \cite{thrun2005probabilistic} and Kalman filters \cite{kalman1960} are classical algorithms that rely on a measurement and transition model for estimating states from sensor data. When a single measurement modality is insufficient, a rich suite of sensor fusion techniques can efficiently fuse multimodal observations (e.g., RGB, LiDAR, radar) into a single state estimate \cite{duan2022multimodal, lee2020multimodal, xu2022review, luo2002multisensor, DBLP:journals/jfr/SpethGRSBMP22}. In our setting, we study world models that \textit{learn} a latent state representation and dynamics model from sequences of high-dimensional observations and robot actions, contrasting the classical settings that assume knowledge of measurement and dynamics models. Because latent world models lack the handcrafted components provided by a domain expert, we refer to ``state estimation'' in our setting as discerning whether the latent representation learned from observations contains sufficient information for predicting downstream outcomes.

\para{Latent World Models} 
A predominant approach for latent world modeling is to train a state representation and dynamics model using reconstruction-based objectives \cite{ha2018world, hafner2019learning} or pretrained foundation model encoders \cite{zhou2024dino}. %
These methods may cause the model to attend to irrelevant details in the observations, motivating the design of world models which operate on more minimal representations via inverse dynamics objectives for learning a model of the agent's dynamics~\cite{efroni2022provably, lamb2022guaranteed} or reward prediction  ~\cite{tian2023toward, hansen2024tdmpc2, huang2022action} to learn a minimal representation suitable for model-based reinforcement learning.  We address a complementary question of whether the observations suitably shape the latent representation to encode safety-relevant quantities to be used for downstream safe control.

\para{Learning with Privileged Information}
Using privileged information available only during training is a widely adopted strategy for learning representations and behavior under partial observability ~\cite{vapnik2009new,hoffman2016learning,pinto2018asymmetric,chen2020learning,kumar2021rma}. Across these settings, additional training-time information guides learning while the deployed robot policy operates with limited observations. Our multimodal supervision follows this general principle to mitigate safety-relevant estimation gaps, using auxiliary sensing modalities during training to shape a latent world model deployed with RGB-only inputs. We study how this supervision improves downstream latent-space safety control.

\section{What Do Latent Safety Controllers Do \textit{Without} Constraint Observability?}
\label{sec:failures-without-obs}

A central assumption of latent-space safe control \cite{nakamura2025generalizing, seo2025uncertainty} is that safety-critical features (i.e., variables needed to model the constraint) are observable in the latent state.
In this section, we first provide a brief background on how safe control is performed in a latent world model, and then demonstrate empirically that violating the constraint observability assumption can result in unsafe policies.

\subsection{Background: Safe Control in Latent World Models}
\label{sec:background}
Let $\mathcal{D} = \{\{(o_t, a_t, l_t)\}_{t=1}^{T} \}_{i=1}^{N}$ be the world model training dataset where $o \in \obsSpace$ is the robot's observations (e.g., RGB images), $\action \in \actionSpace$ are robot actions (e.g., end-effector actions) and $l \in \{ \texttt{safe}, \texttt{unsafe}\}$ are labels of constraint violations. Generally, latent safe control requires: %

\vspace{-0.4cm}
\[ \text{Encoder: }\eqnmarkbox[ocean]{enc}{z = \enc(o)} \quad\quad \text{Latent Dynamics: } \eqnmarkbox[golden_orange]{dynz}{z' \sim \dyn(z,a)} \quad\quad \text{Classifier: } r = \ell(z) \]
Here, the encoder can be frozen and pretrained \cite{zhou2024dino} or learned from scratch alongside the dynamics model %
\cite{hafner2019learning}. 
The safety classifier $\marginfunc(\latent)$ is an additional component trained on top of the world model that outputs a value $r \in \mathbb{R}$ 
such that $r_t \leq 0$ if $l_t = \texttt{unsafe}$ and $r_t \ge 0$ otherwise. 
For safe control, we seek to ensure that $\min_t r_t \geq 0$, indicating that the system never violates the safety constraint. We implement two common safe control methods to safeguard a nominal policy, $a^\text{nom} = \pi^\text{nom}(\obs)$:

\vspace{-0.75cm}
\[
\begin{minipage}{0.46\textwidth}
\[
\begin{aligned}
&\text{\textbf{Least-Restrictive Filtering \cite{bansal2017hamilton, nakamura2025generalizing}:}} \\ 
&Q(\latent,\action) = \min\{r, \max_{\action'} Q(\latent', \action')\}, \\
&a^{\mathrm{safe}}
=
\begin{cases}
\action^{\mathrm{nom}},
 \quad \text{if } Q(\latent, \action^{\mathrm{nom}}) > \epsilon, \\
\displaystyle \action^\shield = \arg\max_{\action \in \actionSpace} Q(\latent,\action),
\quad  \text{else.}
\end{cases}
\end{aligned}
\]
\end{minipage}
\hfill
\begin{minipage}{0.50\textwidth}
\[
\begin{aligned}
&\text{\textbf{Model-Predictive Shielding \cite{bastani2021safe, li2020robust, sinha2023closingloopruntimemonitors}:}} \\[-1pt]
&\latent_{k+1} = \dyn(\latent_k, \action^{\mathrm{nom}}_k),
\qquad k=t,\ldots,t+H, \\ 
&\action^{\mathrm{safe}}_{t:t+H}
=
\begin{cases}
\action^{\mathrm{nom}}_{t:t+H},
& \text{if } \max_{k}~ p_{\mathrm{fail}}(\latent_k) < \epsilon, \\[4pt]
\action^{\mathrm{backup}}_{t:t+H},
& \text{else.}
\end{cases}
\end{aligned}
\]
\end{minipage}
\]
where $\epsilon$ are thresholds tuned for each method, and the MPS controller safeguards $H$-step action chunks from $\policyTask$.
In least-restrictive filtering, we learn a safety critic and safety policy by approximately solving a Hamilton-Jacobi-Bellman equation with latent space reinforcement learning \cite{nakamura2025generalizing}. At runtime, safety filter switches between the nominal policy and the safe policy, $\action^\shield$, depending on how far the system is from the unsafe set (i.e., the set of all states that are assigned negative values by the critic). 
Model-predictive shielding (MPS) foregoes learning an explicit critic and safety policy by taking an online, model predictive control approach. It leverages the (latent) dynamics and assumes a \textit{pre-designed} safe fallback plan, $\action^{\mathrm{backup}}_{t:t+H}$. 
At runtime, MPS predicts the closest to failure the robot could get under the nominal policy for a fixed time horizon. The failure probability $p_{\text{fail}}(z)$ is obtained by treating $r = \ell(\latent)$ as the logit of a binary classifier and mapping it to a probability.

\subsection{Experiment: Estimation and Prediction Gaps in Latent-Space Safe Control}

\para{\textcolor{ocean}{\textit{Estimation Gap:}} \wax}
\label{sec:obs-hardware} To investigate estimation gaps, we design a hardware task where a Franka Research 3 holds a pan full of wax over a hotplate (left, Fig.~\ref{fig:frontfig}). The safety constraint is to not overheat the wax (temperature above  $72^{\circ}$C ). 
The manipulator has access to both a wrist-mounted FLIR Lepton infrared (IR) camera and Intel Realsense D415 RGB camera. 
We train a Dreamer \cite{hafner2023dreamerv3} world model on 180 teleoperated ``lift'' trajectories where the robot's end-effector is constrained to only move along the z-axis with no rotation. 
The IR camera captures temperature ranges between $57$ and $77^{\circ}$C. We automatically label failures whenever the average pixel intensity of the wax region exceeds $0.75$ in the IR image, which corresponds to an average temperature of $72^{\circ}$C.
This often correlates with RGB images where part of the wax has visibly melted. 

\textbf{Latent Safe Control.} We implement a least-restrictive safety filter. Specifically, we train two latent safety filters \cite{nakamura2025generalizing} via reinforcement learning in two world models, one with RGB-only data and the other on multimodal (RGB+IR) data. As described in Sec.~\ref{sec:background}, the filter learns a safety action-value function and a fallback policy. The filter uses observations (RGB or RGB+IR) at each timestep and intervenes with the learned fallback policy if the safety value function predicts that the nominal action will lead to downstream failure. 
The nominal policy $\pi^{\text{nom}}$ keeps the pan at rest on the hotplate.

\textbf{Quantitative Results.} If the safety filter has properly learned to predict and prevent failures, it should lift the pan before the wax overheats, accounting for residual melting after being removed from the hotplate. Across 20 trials, the multimodal safety filter always lifts the pan to prevent overheating; however, the RGB-only filter \textit{fails} to lift the pan and prevent burns 85$\%$ of the time. 

\para{\textcolor{golden_orange}{\textit{Prediction Gap:}} \rice} To investigate prediction gaps, we design a hardware task where a manipulator executes a pouring motion above a plate while holding an opaque bottle that may be completely empty or filled with rice (right, Fig.~\ref{fig:frontfig}). The safety constraint is to prevent rice from spilling onto the plate.
The manipulator has access to both a wrist-mounted Zed Mini RGB camera and a GelSight Mini attached to its finger.
For world modeling, we collect 650 trajectories of various pouring and tilting motions, where the bottle is empty in half of the trajectories. We train a ViT-based world model in the style of DINO-WM \cite{zhou2024dino} with additional tactile conditioning. We label failures whenever $>5$ grains of rice are visibly spilled out of the bottle in the RGB image. This ensures that the RGB modality can identify failures, but cannot see if the bottle is filled before spilling.

\textbf{Latent Safe Control.} 
We follow the model-predictive shielding formulation from Sec.~\ref{sec:background}. 
To focus our analysis on the world model and failure classifier's understanding of future outcomes, we deploy the an unfiltered controller ($\epsilon = 1$) and record the predicted failure probability for the action chunk that induces a failure. %
We deploy both a RGB-only and RGB+Tactile world model (and corresponding failure classifiers).
Our $\pi^{\text{nom}}(o)$ is a diffusion policy~\cite{chi2024diffusionpolicy} that takes RGB images as input and is trained to spill the bottle. The world model simulates a $16$-step action chunk from $\pi^{\text{nom}}$, and we record the maximum predicted failure probability over the chunk. We then execute the first $8$ steps before requerying the policy.

\textbf{Quantitative Results.} 
We run 50 trials of the robot interacting with a filled bottle of rice. The RGB-only WM is overly optimistic: it predicts that a spill is \textit{unlikely} (i.e.,  $p_\text{fail}(\latent) < 0.2$) on over $50\%$ of the action chunks that ultimately \textit{do cause a spill}. In contrast, the multimodal WM is more calibrated: of all the spill-inducing action sequences, it predicts that a spill is unlikely for only one.

\section{How Can We Quantify Constraint Observability in  Latent Spaces?}
\label{sec:quantifying-obs}

As shown in Sec.~\ref{sec:failures-without-obs}, the latent safe control policies can exhibit degenerate behaviors when safety-relevant features are difficult to observe. 
Because our safe controllers act in the world model's learned latent-space, it is critical to understand how partial observability affects what the world model latent state can represent. We propose two diagnostics for understanding constraint observability in the learned latent state: (1) an observation-level mutual-information estimate that quantifies whether that latent state contains safety-critical information at all, and (2) an open-loop rollout metric for identifying when world models can fail to predict downstream safety outcomes.

\para{Quantifying Observability via Mutual Information}
To assess whether safety-relevant quantities are observable from an observation, such as an RGB image, we propose estimating the mutual information between the observation $X$ and the safety label $Y$~\cite{shannon}. For random variables $X$ and $Y$ drawn from a joint distribution $P(X,Y)$, we can define their mutual information as
\begin{equation}
    I(Y;X) = H(Y) - H(Y \mid X),
\end{equation}
where $H(Y)$ is the entropy of the safety label and $H(Y \mid X)$ is its conditional entropy given the observation. Because directly estimating mutual information is difficult in general~\cite{kraskov2004estimating}, we use the Barber--Agakov variational lower bound~\cite{barberagakov}:
\begin{equation}
    I(Y;X) \ge H(Y) - H(P,Q),
    \qquad
    H(P,Q) = -\mathbb{E}_{P(X,Y)} \log Q(Y\mid X),
\end{equation}
where $P(Y \mid X)$ is the true conditional distribution and $Q(Y \mid X)$ is a variational estimate.

In our safety setting, it is efficient to generate empirical estimates of $H(Y)$ and $H(P , Q)$ since the safety label $Y$ is a binary random variable. 
For any latent world model, we take the encoder of the world model that maps $\obsSpace \rightarrow \latentSpace$ 
and train a linear probe \cite{alain2016understanding} on top of the resulting embeddings. 
The linear probe is trained to output the logits of a binary random variable via binary cross-entropy loss, and its logits are calibrated using temperature scaling \cite{guo2017calibration} on the calibration dataset. 
We estimate the cross-entropy between the empirical data distribution, $\hat{P}(Y|X)$, and the predictor, $Q(Y|X)$, to obtain $H(\hat{P}, Q)$, and estimate $H(Y)$ directly from the dataset distribution%
, $\hat{P}(Y)$. 
This estimates a lower bound on how much the uncertainty over the safety label is reduced by an observation.

\para{Experimental Setup} To obtain our estimate of mutual information, we use trajectories unseen by the world model encoders to train, calibrate, and evaluate the linear probe. 
We use 40 trajectories for \wax and 650 trajectories for \rice, and divide each dataset into an 80/10/10 split for training, calibration, and evaluation, respectively.

\begin{wraptable}{r}{0.56\textwidth}
\vspace{-0.65cm}
\centering
\setlength{\tabcolsep}{3pt}
\small
\begin{tabular}{l|ccc|ccc}
\toprule
& \multicolumn{3}{c|}{\wax} 
& \multicolumn{3}{c}{\rice} \\
Metric & RGB & IR & MM & RGB & Tactile & MM\\
\midrule
$\frac{I(Y;X)}{H(Y)}$ & 0.221 & 0.801 & 0.853 & 0.962 & 0.221 & 0.968  \\
Acc. & 0.870 & 0.969 & 0.979 & 0.994 & 0.838 & 0.995 \\
B. Acc. & 0.626 & 0.983 & 0.988 & 0.988 & 0.687 & 0.988 \\
\bottomrule
\end{tabular}
\caption{\textbf{Safety-Relevant Signal in Observations.} Mutual information estimates and classification accuracies.}
\label{tab:mutualinf}
\vspace{-0.8cm}
\end{wraptable}

\paranopunc{Q1: How much do different observation modalities reveal safety-critical information?}

We train multimodal world models for both tasks (RGB+IR for \wax and RGB+Tactile for \rice), along with a linear probe on top of the encoded RGB image $e_{\text{RGB}}$, the non-RGB modality (e.g., IR or tactile) $e_{\text{aux}}$, and their composition, $e_{\text{RGB}} \oplus e_{\text{aux}}$.
In Table~\ref{tab:mutualinf}, we report (i) our \textit{mutual information (MI) estimate} between the embedding and the safety label, normalized by its upper bound $H(Y)$, (ii) \textit{classification accuracy} of the linear probe, and (iii) \textit{balanced accuracy}. Note that all metrics are between 0 and 1, enabling fair comparisons.

\para{Results} In the \wax~task, which is designed to expose an estimation gap, the normalized MI estimate separates RGB and IR more sharply than the accuracy-based metrics: RGB achieves $I(Y;X)/H(Y)=0.221$, while IR achieves $0.801$, a much larger gap than either accuracy-based metric. This suggests that accuracy can overstate how much a given observation modality can detect safety-relevant quantities. In the \rice~task, the gap between RGB and multimodal observations is smaller since the RGB modality is sufficient to detect failures once they occur.

\paranopunc{Q2: How well can safety outcomes be predicted from world model rollouts under partial observability?}
Even if the latent state space is sufficient for identifying the current failure label, it may lack the representational capability to \textit{predict} future safety consequences. 
To study this, we compute the failure classification accuracy when rolling out the world model open-loop for 16 timesteps, after providing 3 timesteps of history from a held out evaluation set. 

\begin{wrapfigure}{r}{0.5\linewidth}
    \vspace{-1.3cm}
    \centering
    \includegraphics[width=\linewidth]{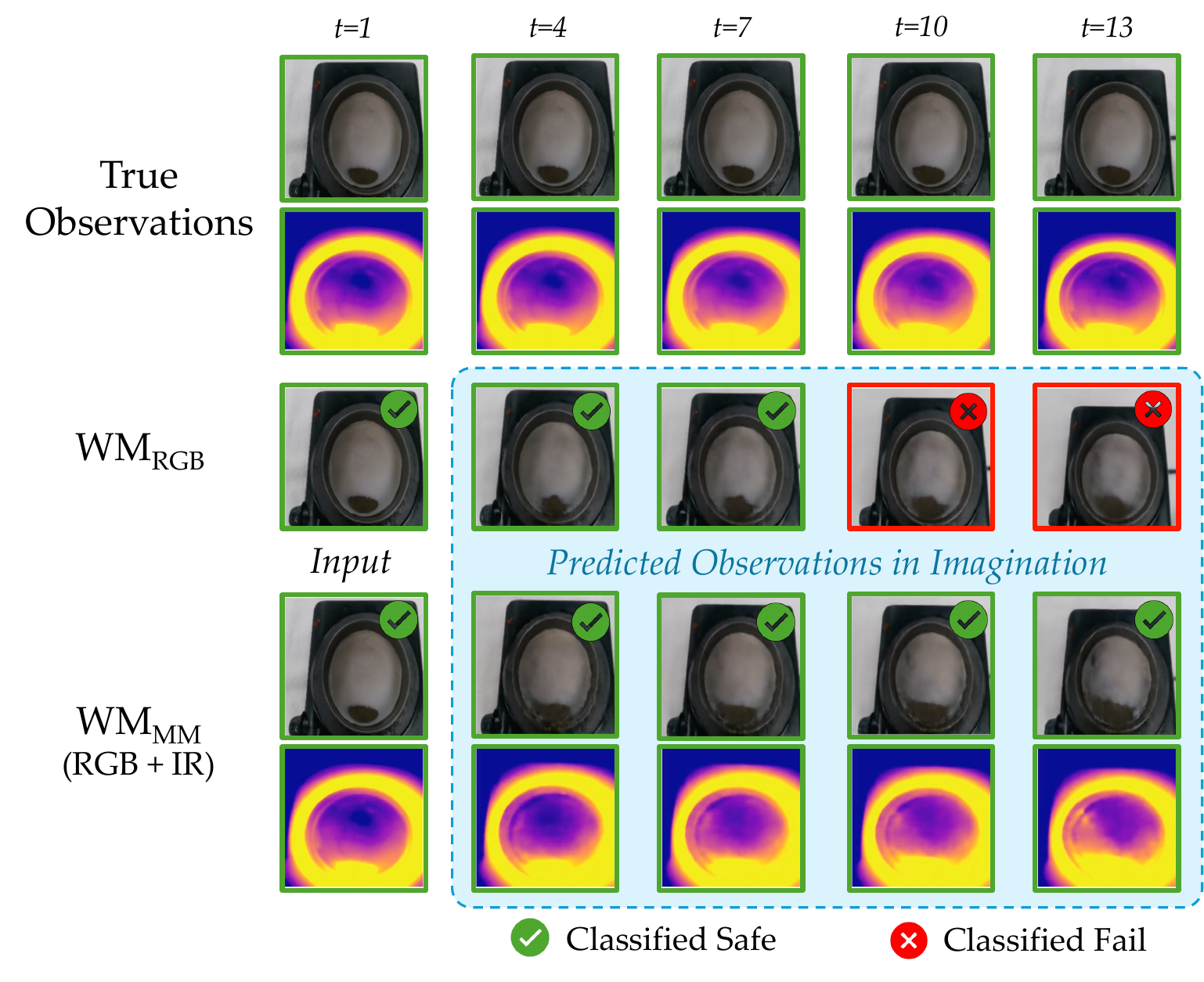}
    \vspace{-2em}
    \caption{\textbf{Real vs. Imagined Observations \& Safety Labels.
    } Open-loop predictions of a held-out subtrajectory reveal that \wmrgb~ misjudges whether lifting the wax plate will cause failure. In contrast, observing temperature via IR images allows the \wmmm~to correctly classify that this action leads to a safe outcome.}
    \vspace{-0.8cm}
    \label{fig:hw_ol}
\end{wrapfigure}

\para{Results} We report our results in Table~\ref{tab:ol_imag}. 
While the aggregate classification accuracy is high for all settings and methods, we find that the multimodal world model has significantly higher accuracy on samples where both world models disagree. %
To further investigate the nature of any misclassifications, we visualize, in Figure \ref{fig:hw_ol}, a %
trajectory where the robot actively lifts the wax plate. We compare the true observations to the open-loop  world model's imaginations, given the action sequence of lifting the pan for both the RGB and MM world model.  We notice that while both models are able to predict visual changes, \wmrgb~falsely imagines that this action sequence leads to a safety violation. In other words, the latent representation of \wmrgb~\textit{has a weaker understanding of how actions correlate with downstream safety outcomes}.

\begin{table}[h!]
\centering
\setlength{\tabcolsep}{6pt}

\begin{minipage}[t]{0.55\linewidth}
\centering
\begin{tabular}{@{}l|cccc|c@{}}
\toprule
\textbf{Method}
& \textbf{TS} $\uparrow$
& \textbf{FS} $\downarrow$
& \textbf{FU} $\downarrow$
& \textbf{TU} $\uparrow$
& \textbf{F1} $\uparrow$ \\
\midrule
Wax RGB  & 0.751 & 0.018 & 0.149 & 0.083 & 0.899 \\
Wax MM   & 0.896 & 0.020 & 0.004 & 0.080 & \textbf{0.986} \\
\midrule
Rice RGB & 0.750 & 0.015 & 0.003 & 0.232 & 0.988 \\
Rice MM  & 0.752 & 0.014 & 0.000 & 0.233 & \textbf{0.991} \\
\bottomrule
\end{tabular}
\caption*{\textbf{(a) Aggregate outcomes}}
\end{minipage}\hfill
\begin{minipage}[t]{0.43\linewidth}
\centering
\begin{tabular}{@{}cccc|c@{}}
\toprule
\textbf{TS} $\uparrow$
& \textbf{FS} $\downarrow$
& \textbf{FU} $\downarrow$
& \textbf{TU} $\uparrow$
& \textbf{F1} $\uparrow$ \\
\midrule
0.015 & 0.034 & 0.901 & 0.050 & 0.031 \\
0.901 & 0.050 & 0.015 & 0.034 & \textbf{0.965} \\
\midrule
0.000 & 0.391 & 0.356 & 0.253 & 0.000 \\
0.356 & 0.253 & 0.000 & 0.391 & \textbf{0.738} \\
\bottomrule
\end{tabular}
\caption*{\textbf{(b) Disagreements only}}
\end{minipage}

\caption{\textbf{Safety Outcome Prediction.} Outcomes of 16-step open-loop trajectories. The right table reports results only on samples where RGB and MM disagree on safety outcomes; rows follow the same method ordering as the left table.}
\label{tab:ol_imag}
\vspace{-1.5em}
\end{table}

\section{Mitigating Partial Observability in RGB-only Latent Space Control}
\label{sec:multimodal-training}

So far, we showed that estimation and prediction gaps appear in latent world models when the set of training observations is insufficient. We also introduced mutual information and rollout prediction accuracy as diagnostic tools to quantify this gap. We now ask whether these diagnostics predict the behavior of deployed safety controllers, and what interventions help mitigate the problems introduced by partial observability.

\subsection{\wax: Privileged Supervision Mitigates  Estimation Gaps}
In Sec.~\ref{sec:quantifying-obs}, we diagnosed that estimation gaps can arise when uninformative observations (e.g., RGB-only) induce latent states that fail to capture safety-relevant quantities (e.g., temperature). A natural solution is to equip both the world model and the robot with additional sensing modalities (e.g., RGB and IR). However, we hypothesize that such richer sensing is not necessary at deployment, provided the model learns to extract the right information from unimodal observations. 
Here, additional modalities can play a supervisory role during training, guiding what the model should attend to. Specifically, by requiring the world model to predict an auxiliary modality that is correlated with safety-relevant properties inferred from the test-time input, we can shape the latent representation toward a more complete model of the underlying state of the world.

\para{Method} We introduce a world model training strategy that uses RGB inputs but \textit{provides supervision by forcing the latent to reconstruct multimodal data}: both RGB and IR images. We refer to this as \wmmasked.
Note that this is different from the multimodal world model in Sec.~\ref{sec:quantifying-obs} %
that receives both RGB and IR images input into its encoder.

\para{Results: Mutual Information \& Open-Loop Prediction Accuracy} We first measure the normalized MI using the \wmmasked~encoder to be $\frac{I(Y;X)}{H(Y)} = 0.15$, which is aligned with the results for \wmrgb. This is expected, since our multimodal reconstruction will not fundamentally add more information to the RGB input modality. %
However, the open-loop predictions of \wmmasked~have an $F_1$ score of $0.943$ which closely matches the \wmmm~baseline of $0.986$. 
While our experiments cannot isolate whether the benefit arises from stronger latent representations or improved dynamics predictions, our results suggest that jointly supervising future RGB and IR predictions enables the world model to better predict how actions influence downstream safety consequences.

\para{Setup: Closed-loop Evaluation \& Metrics} Next, we look at how the resulting safety controller behaves with and without multimodal supervision. During deployment, we use a simple nominal policy, $\policyTask$, that grips a pan full of wax and holds it fixed over the hot plate. 
We deploy three latent safety filters computed in three corresponding world models: RGB-input and supervision (\wmrgb), multimodal (RGB+IR) input and supervision (\wmmm), and one with RGB inputs but multimodal supervision (\wmmasked). 
Each filter is deployed in a least-restrictive fashion as in Sec.~\ref{sec:background} with a fixed $\epsilon = 0.3$ as our threshold for all deployments. 
We stop each deployment trajectory after a safe lift or after the learned $\marginfunc(\latent)$ falls below zero. 
We deploy each filter 20 times and report the safety performance in Table~\ref{tab:safety}. 
We report the \textit{lift rate}, which measures whether or not the fallback policy (i.e., $a^\shield = \arg\max_\action Q(\latent, \action)$) is able to learn a non-myopic recovery behavior, and the \textit{average pixel intensity} at intervention, measuring how close to failure the system is during filter intervention. %

\begin{wraptable}{r}{0.65\textwidth}
\vspace{-0.5cm}
\centering
\setlength{\tabcolsep}{3pt}
\renewcommand{\arraystretch}{1.3}
\small
\begin{tabular}{lccc}
\toprule
& \multicolumn{3}{c}{\textbf{World Models \& Respective Safety Filters}} \\
\cmidrule(lr){2-4}
\textbf{Metric} & \wmrgb & \wmmm & \wmmasked \\
\midrule

\makecell[l]{Lift Rate \\ {[95\% CI]}}
& $0.15\,[0.04,\,0.39]$
& $1.0\,[0.80,\,1.0]$
& $1.0\,[0.80,\,1.0]$ \\

\makecell[l]{Pixel Intensity \\ at Intervention}
& $0.587 \pm 0.303$
& $0.695 \pm 0.073$
& $0.522 \pm 0.208$ \\

\bottomrule
\end{tabular}

\caption{\textbf{Hardware: Closed-Loop Safety Filter Performance.}
Lift rates over 20 trials with 95\% Wilson-CC confidence intervals~\cite{newcombe1998}. \wmrgb{} fails to lift reliably. \wmmm{} intervenes before failure, while \wmmasked{} intervenes more variably, likely because it cannot directly observe safety-relevant quantities.}
\label{tab:safety}

\vspace{-0.4cm}
\end{wraptable}

\begin{figure}[tbh]
    \centering
    \includegraphics[width=\linewidth]{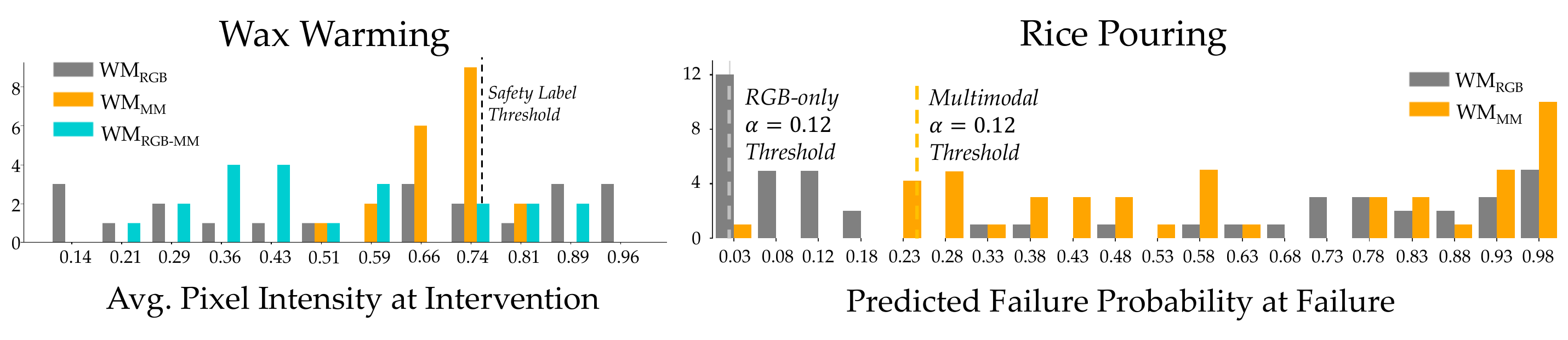}
    \caption{\textit{(Left)} Ground-truth pixel intensities when each safety filter intervenes. The \wmmm~policy that takes in IR images concentrates just prior to the threshold used for failure labeling, whereas \wmrgb~and \wmmasked~use only RGB as input are spread more conservatively, indicating they are unable to discern \textit{when} to filter.
    \textit{(Right)} Distribution of predicted failure probabilities during conformal calibration. The \wmrgb~frequently fails to predict spills leading to conservative thresholds.}
    \label{fig:histograms}
\end{figure}

\para{Results: Multimodal-Supervised Latent Safety Control} We find that \wmmasked~now learns a safety fallback policy that lifts the pan in $100\%$ of the trials to prevent overheating. Interestingly, while the RGB input still does not provide any information on \textit{when} to intervene, we find that \wmmasked~is \textit{conservative}, rather than optimistic, compared to \wmmm. We visualize this conservativeness in Fig.~\ref{fig:histograms} by plotting the distribution of the temperature at intervention. While \wmmm matches a least-restrictive filter and intervenes close to the failure boundary, \wmrgb~and \wmmasked~intervene earlier.

An important implication of this result is that, while a latent safety filter may learn that it \textit{should} intervene, this does not imply that the safety policy knows \textit{how} to intervene.
The safety policy learned by \wmrgb~leaves the pan on the hotplate $85\%$ of the time, whereas \wmmasked~, trained to reconstruct RGB and IR observations from RGB-only inputs, learns a policy that lifts the pan in all 20 trials.
These results are consistent with multimodal supervision helping the world model capture the relative safety consequences of actions: lifting the pan reduces overheating risk compared with leaving it in place.
However, despite matching the $100\%$ lift rate of \wmmm~, \wmmasked~intervenes earlier and at more variable temperatures (Fig.~\ref{fig:histograms}, Left).
This suggests that learning how to intervene effectively does not imply accurately estimating distance to the safety boundary.

\subsection{\rice: Calibrating Failure Classifiers Can Mitigate Prediction Gaps}
\label{sec:tac}
For prediction gaps, partial observability does not necessarily prevent the WM from representing safety failures after they occur, but it prevents it from reliably forecasting them. 
To see this, we isolate the effect in the Rice Pouring task and tilt a filled bottle with a diffusion policy 50 times, and evaluate the WM's predicted failure probability for the $H$-step action chunks that lead to a spill. 

We visualize the resulting prediction distributions in Figure~\ref{fig:histograms}. We find that the RGB-only world model is capable of representing spills, but its predictions are miscalibrated: it assigns lower failure probabilities than the multimodal model to many trajectories that in fact result in spilling, and remains overly-optimistic. In other words, although the model can express the possibility of failure, it cannot align predicted probabilities with true downstream risk under partial observability.

\para{Method} 
We address prediction gaps not by changing the WM representation, but by calibrating the MPS shielding threshold. 
In particular, we apply a thresholding procedure that adjusts the $\epsilon$ threshold in the MPS paradigm from Sec.~\ref{sec:background} that controls the false-safe rate, i.e., the rate at which the nominal policy is permitted to spill the bottle.
We choose this threshold using conformal prediction~\cite{conformal} by treating the 50 collected spill trajectories as calibration data. We perform trajectory-level calibration~\cite{knowno2023, seo2025uncertainty}, and set the target false-safe rate to $\alpha = 0.12$ and choose $\epsilon$ as the empirical $\alpha$-quantile of the predicted failure probabilities on unsafe calibration trajectories with a finite sample correction, yielding $\epsilon = 0.02$ for the RGB MPS controller and $\epsilon = 0.24$ for the multimodal controller. Additional details on the conformal calibration are in the Appendix~\ref{app:conformal}. At deployment, a candidate action sequence is accepted only if its predicted failure probability is below the calibrated threshold; otherwise, control switches to the fallback policy. Under the standard assumption that calibration and test trajectories are exchangeable, this bounds the expected false acceptance rate at approximately $\alpha$.

\begin{wraptable}{r}{0.55\textwidth}
\centering
\setlength{\tabcolsep}{2pt}
\begin{tabular}{llcc}
\toprule
Bottle & Model & Pour Motion & Fallback  \\
& & [95\% CI] & [95\% CI] \\
\midrule
\multirow{2}{*}{Filled} 
& RGB & 0.14 [0.06, 0.27] & 0.86 [0.73, 0.94] \\
& MM  & 0.14 [0.06, 0.27] & 0.86 [0.73, 0.94] \\
\midrule
\multirow{2}{*}{Empty}  
& RGB & 0.18 [0.09, 0.32] & 0.82 [0.68, 0.91] \\
& MM  & 0.60 [0.45, 0.73] & 0.40 [0.27, 0.55] \\
\bottomrule
\end{tabular}
\caption{\textbf{Results: MPS for 50 trials.} Rates are reported with 95\% Wilson-CC confidence intervals~\cite{newcombe1998}. After conformal calibration to control false acceptance, the multimodal MPS controller retains greater control authority in the empty-bottle setting while maintaining fallback behavior in higher-risk settings.}
\label{tab:rice_metrics}
\end{wraptable}

\para{Evaluation \& Metrics} 
We roll out the calibrated MPS controller for each world model 50 times where the bottle is filled and 50 times where the bottle is empty.
We stop each deployment trajectory after a spill or when the MPS controller overrides the base diffusion policy.
For the filled bottle trajectories, we use the same failure labeling scheme as the training data and declare the spill motion as being executed only if there were more than 5 grains of rice spilled onto the table. For the empty bottle, we consider the pouring motion executed if the maximum tilt of the end effector exceeds $23^\circ$, which closely resembles the end-effector tilt for spill trajectories.

\para{Results: Calibrated Failure Prediction Mitigates Over-Optimism} 
We report the execution and fallback rates of the pour-motion in Table~\ref{tab:rice_metrics}. For filled-bottle trajectories, both filters fail at a rate close to the target calibration level $\alpha=0.12$. However, the multimodal world model is less conservative under MPS with statistical significance, as its latent-state predictions more accurately identify safe trajectories and allow pouring motions for the empty bottle at a rate of $60\%$ compared to $18\%$ for the RGB-only filter. %
This shows that when the world model is unable to consistently predict spills, our method leads to a \textit{pessimistic} safe controller. We note that this is a standard tradeoff in safety-critical control where it is common to adopt more conservative behavior when predictive models are uncertain or untrustworthy~\cite{coffee, fisac2019general}.

\section{Conclusion \& Limitations}

We study how partial observability degrades latent world models and their controllers. Through controlled hardware experiments,  we identify two failure modes: estimation gaps, where latent states cannot determine current safety, and prediction gaps, where they can detect failures but not predict their onset.
We introduce quantitative diagnostics for both, along with mitigations: multimodal supervision for estimation gaps and conformal calibration for prediction gaps for safer control under RGB-only deployments.
Overall, our results show that accurate RGB prediction is not enough: safe control requires latent representations that retain the information needed to reason about safety.

\para{Limitations} 
While we use mutual information to quantify when latent states fail to capture safety, how best to inject the missing safety information remains unclear. Our experiments are limited to controlled domains where observability can be precisely manipulated; in more complex environments, identifying when latent spaces fail to encode safety is harder. Finally, although our mitigation strategies enable RGB-only control in our case studies, more general solutions are needed for partial observability in world models. These limitations point to promising directions for future work.

\section{Acknowledgements}

MK and KN would like to thank Steven Man for his assistance with 3D printing and designing custom camera mounts for our experiments. The authors would also thank Junwon Seo, Gokul Swamy, Ran Tian, and Rosy Chen for insightful discussions and valuable feedback throughout this work. KN would additionally like to thank Yilin Wu for assistance during the tactile experiments. KN was supported by the National Science Foundation Graduate Research Fellowship Program under Grant No. DGE2140739.  This work was funded in part by the NSF CAREER Award No. 2441014 and the DARPA Young Faculty Award (YFA). Any opinions, findings, conclusions or recommendations expressed in this material are those of the author(s) and do not necessarily reflect the views of the National Science Foundation, DARPA, or the U.S. Government.

\bibliography{references}
\section{Appendix}

\subsection{Conformal Prediction}
\label{app:conformal}

We briefly describe conformal prediction \cite{conformal} and review the details of our threshold calibration procedure in Section \ref{sec:tac}. Conformal prediction is a statistically principled framework for constructing prediction sets that contain the true class label with user-specified probability $1-\alpha$. Its guarantees rely only on mild distributional assumptions, such as exchangeability or i.i.d. sampling of the calibration and test data.

More concretely, suppose $\{(X_i,Y_i)\}_{i=1}^n$ and $(X_{\mathrm{test}},Y_{\mathrm{test}})$ are exchangeable input-output pairs, and define a nonconformity score $s(x,y)$ such that larger values indicate worse agreement between $x$ and $y$. Let $\hat{q}$ be the $\lceil (n+1)(1-\alpha) \rceil / n$ empirical quantile of the calibration nonconformity scores. Then the prediction set
\[
C(X_{\mathrm{test}}) = \{y : s(X_{\mathrm{test}}, y) \leq \hat{q}\}
\]
satisfies
\[
p(Y_{\mathrm{test}} \in C(X_{\mathrm{test}})) \geq 1-\alpha.
\]

In our model-predictive shielding controller, we want the fallback controller to be triggered whenever an action chunk from the nominal diffusion policy $\pi_{\mathrm{nom}}$ is likely to spill the bottle. For each unsafe trajectory in our calibration dataset, we identify the executed $H$-step action chunk that led to a spill. We define the nonconformity score for this trajectory as
\[
s = 1 - \max_k p_{\mathrm{fail}}(z_k),
\]
where $\max_k p_{\mathrm{fail}}(z_k)$ denotes the worst-case predicted probability of a spill over the world-model rollout induced by the action chunk $a_{t:t+H}$. A large value of $s$ means that the WM assigned low failure probability to an action chunk that in fact caused a spill.

Taking $\hat{q}$ to be the $\lceil (n+1)(1-\alpha) \rceil / n$ empirical quantile of these scores gives the intervention threshold
\[
\epsilon = 1 - \hat{q}.
\]
At deployment, the model-predictive shield accepts a candidate action chunk only if
$\max_k p_{\mathrm{fail}}(z_k) < \epsilon$
otherwise, it switches to the fallback controller.

We can rewrite the conformal prediction guarantee and observe 
\[p\big(\max_k p_{fail}(z_k) < \epsilon \;|\; a_{t:t+H} \text{ leads to spill}\big) \leq  \alpha,  \]
meaning that under the assumption of exchangeability between trajectories, this ensures that the probability of accepting a spill-inducing action chunk is at most $\alpha$.

\subsection{Illustrative Example: Thermal Unicycle}
\label{sec:obs-unicycle}

The Thermal Unicycle is a controlled analogue of the \wax~hardware experiment. Its purpose is to isolate an estimation observability gap in a setting where the ground-truth safety-relevant variable is known. The vehicle corresponds to the wax pan, the circular hot region corresponds to the hot plate, and the scalar heat state corresponds to wax temperature. By constructing RGB observations that only reveal overheating after the vehicle leaves the hot region, while IR observations directly reveal heat, this example lets us test whether a latent safety controller fails when the relevant safety variable is absent from the learned observation space.

\begin{figure}[h]
    \centering
    \includegraphics[width=0.99\linewidth]{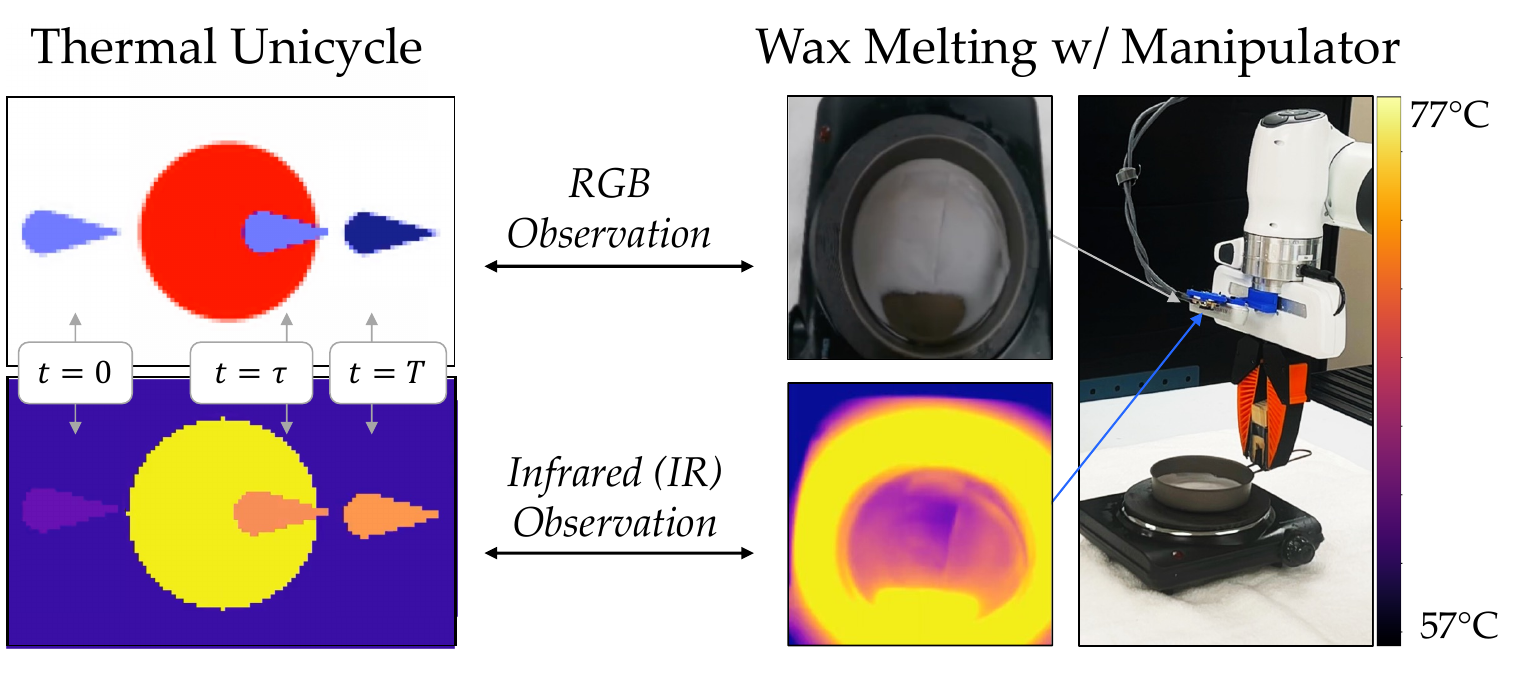}
    \caption{\textbf{Experiment Testbeds.} In simulation (left) and \wax~(right), we collect paired RGB and infrared (IR) observations. In both settings, the safety-relevant state variable is heat, which is directly observable from IR but only partially observable from RGB.}
    \label{fig:testbeds}
\end{figure}

\subsubsection{Ground-Truth Dynamics}

We simulate a planar vehicle that can navigate into a ``hot plate'' region, shown in Fig.~\ref{fig:testbeds}. The vehicle is allowed to warm up, but it violates the safety constraint if its heat exceeds a threshold.

We model the ground-truth system as a unicycle augmented with a scalar heat state. The state is
\[
    x = [p_x, p_y, \theta, v, h],
\]
where $(p_x,p_y)$ is position, $\theta$ is heading, $v$ is velocity, and $h \in [0,1]$ is heat. The dynamics are
\[
    x_{t+1}
    =
    x_t
    +
    \Delta t
    \left[
        v_t \cos \theta_t,\;
        v_t \sin \theta_t,\;
        \omega_t,\;
        a_t,\;
        f_h(p_x,p_y)
    \right].
\]
The heat dynamics are
\[
    f_h(p_x,p_y)
    =
    \mathbb{I}\{p_x^2 + p_y^2 < R^2\}\alpha
    -
    \mathbb{I}\{p_x^2 + p_y^2 \geq R^2\}\beta,
\]
where $\alpha=0.03$, $\beta=0.06$, and the hot region is a disk of radius $R=0.5$ centered at the origin. The control input is $u := (\omega,a)$, where $\omega \in [-1.1,1.1]$ is the turn rate and $a \in [-2,2]$ is the linear acceleration. The velocity is bounded by $v \in [0,1]$, and the timestep is $\Delta t = 0.05$. We label the vehicle as unsafe when the ground-truth heat exceeds the threshold $h > 0.8$.

\subsubsection{Observation Models}

Each trajectory produces two image streams: RGB and IR, each with resolution $128 \times 128$. The IR observation is designed to make the safety variable directly observable: the vehicle intensity is a monotone function of heat, so hotter states appear brighter. This provides sufficient information to determine the ground-truth safety label.

The RGB observation is designed to provide partial observability into the true system state. While the vehicle remains inside the hot region, RGB does not reveal its current heat. Instead, overheating becomes visible only after the vehicle exits the hot region, analogous to only seeing that wax has melted after the temperature has already become safety-relevant. Once the vehicle appears burnt in RGB, decreasing heat in later timesteps does not remove this visual evidence. By this construction, RGB contains delayed evidence correlated with failure, but is insufficient for determining the current safety state in isolation. 

This observation construction mirrors \wax: RGB can contain visual evidence of melting, but IR directly measures the underlying thermal state. The simulation lets us test whether latent safety filters trained from partially observable inputs learn to avoid the true safety violation or just \textit{avoid observing} failure.

\subsubsection{WM and Safety Filter Training}

We generate 4{,}000 observation-action trajectories by sampling initial states uniformly within the state bounds and applying random controls sampled from their admissible ranges. Each trajectory is rolled out for at most 100 timesteps, or until the vehicle exits the valid state space. Both RGB and IR observations are recorded for every trajectory.

We train two latent WMs on this dataset. The RGB-only WM, \wmrgb, receives RGB observations only. The multimodal WM, \wmmm, receives both RGB and IR observations. For each WM, we train a least-restrictive latent safety filter following Sec.~\ref{sec:failures-without-obs}. Each filter learns a latent safety critic and fallback policy using rollouts in the learned latent dynamics. At evaluation time, the vehicle is initialized inside the hot region with low heat and zero velocity. A successful safety filter must choose actions that leave the hot region before $h$ exceeds the failure threshold. Specific hyperparameter details can be found in Table~\ref{tab:rssm_hyperparams}.

\subsubsection{Qualitative and Quantitative Results}

\begin{figure}[ht]
    \centering
    \includegraphics[width=0.9\linewidth]{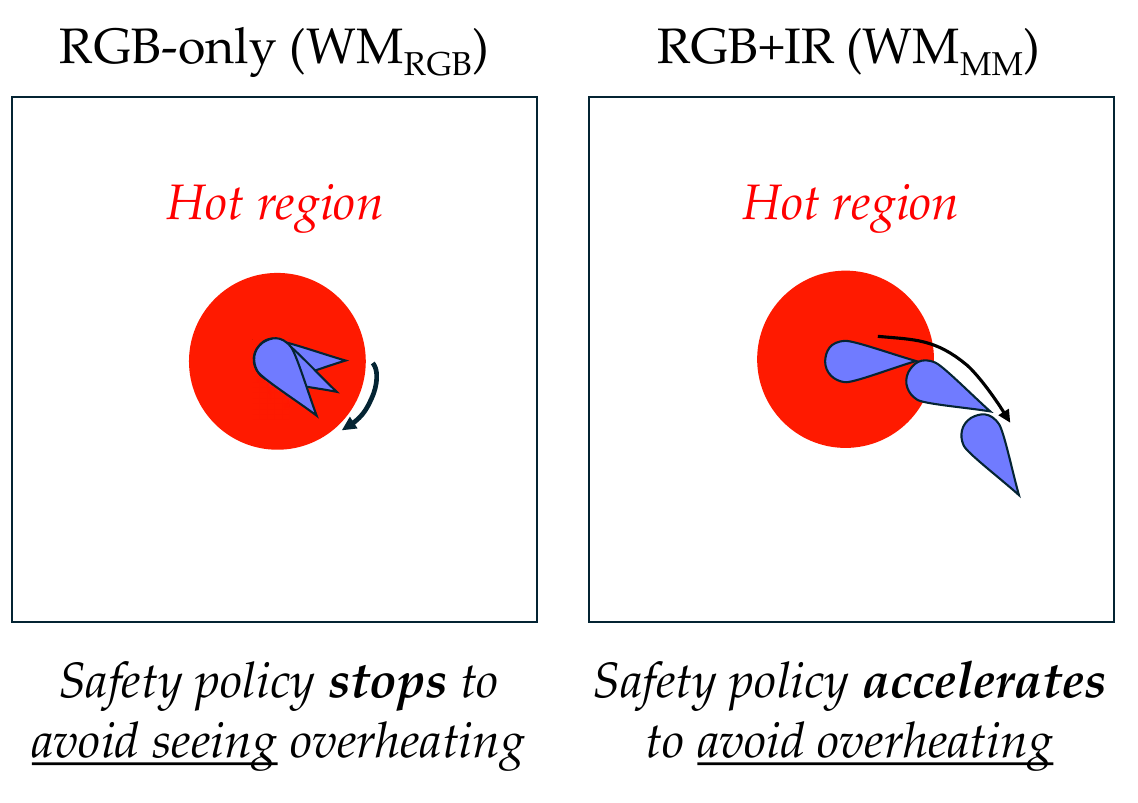}
    \caption{\textbf{Safety Controllers without \& with Observability.} In the \textit{Thermal Unicycle} example, the robot only appears burnt in RGB after it leaves the hot plate. \textit{Left:} The RGB-only safety policy is myopic: it stays in the hot region, avoiding \textit{seeing} failure. \textit{Right:} When the WM is trained with RGB+IR observations, the resulting safety controller accelerates out of the hot region to prevent overheating.}
    \label{fig:dubins}
\end{figure}

Figure~\ref{fig:dubins} shows the resulting closed-loop behavior. The multimodal safety filter trained with RGB and IR accelerates out of the hot region before overheating. Because IR directly reveals heat, the learned latent representation supports a safety critic that correctly assigns high risk to remaining on the hot plate and low risk to escaping.

In contrast, the RGB-only safety filter exhibits a myopic failure mode. Since RGB reveals overheating only after the vehicle exits the hot region, the RGB-only WM and safety classifier can identify failure only after the failure has become visually apparent. The resulting safety policy therefore learns to remain inside the hot region, which avoids observing visual evidence of failure but does not prevent the ground-truth heat violation. This illustrates the central failure mode studied in the paper: under partial observability, a latent safety controller may optimize for safety in a latent space that does not represent the true safety-relevant state.

As in our hardware experiments, we quantify the observability gap at three levels: observation, latent state, and open-loop latent predictions. First, we estimate the normalized mutual information lower bound, $I(Y;X)/H(Y)$, between each observation modality and the safety label. Second, we evaluate whether the learned latent state after $N=5$ history steps correctly classifies the current safety label. Third, we evaluate whether 16-step open-loop rollouts in the WM correctly predict future safety outcomes.

\begin{table}[h]
\centering
\setlength{\tabcolsep}{8pt}
\renewcommand{\arraystretch}{1.12}
\begin{tabular}{l|ccc}
\toprule
\textbf{Modality}
& $\mathbf{I(Y;X)/H(Y)} \uparrow$
& \textbf{Acc.} $\uparrow$
& \textbf{B. Acc.} $\uparrow$ \\
\midrule
RGB    & 0.182 & 0.864 & 0.738 \\
IR     & 0.909 & 0.982 & 0.966 \\
RGB+IR & 0.903 & 0.981 & 0.976 \\
\bottomrule
\end{tabular}
\vspace{0.5em}
\caption{\textbf{Simulation Observation Signal Metrics.} Observation-level safety signal metrics for the simulated Thermal Unicycle task.}
\label{tab:sim_observation_metrics}
\end{table}

\begin{table}[h]
\centering
\setlength{\tabcolsep}{7pt}
\renewcommand{\arraystretch}{1.12}
\begin{tabular}{ll|ccccc}
\toprule
\textbf{Eval.} & \textbf{Method}
& \textbf{TS} $\uparrow$
& \textbf{FS} $\downarrow$
& \textbf{FU} $\downarrow$
& \textbf{TU} $\uparrow$
& $\mathbf{F_1} \uparrow$ \\
\midrule
\multirow{2}{*}{Latent State}
& RGB    & 0.623 & 0.013 & 0.104 & 0.272 & 0.914 \\
& RGB+IR & 0.701 & 0.002 & 0.026 & 0.283 & 0.981 \\
\midrule
\multirow{2}{*}{Open-loop Pred.}
& RGB    & 0.643 & 0.012 & 0.089 & 0.260 & 0.927 \\
& RGB+IR & 0.693 & 0.012 & 0.035 & 0.264 & 0.967 \\
\bottomrule
\end{tabular}
\vspace{0.5em}
\caption{\textbf{Simulation Safety Classification Metrics.} Latent-state and 16-step open-loop prediction metrics for the simulated Thermal Unicycle task.}
\label{tab:sim_classification_metrics}
\end{table}

The observation-level metrics in Table~\ref{tab:sim_observation_metrics} show that IR and RGB+IR contain substantially more safety-relevant information than RGB alone. The latent-state and open-loop prediction metrics in Table~\ref{tab:sim_classification_metrics} show the same trend: adding IR reduces false unsafe predictions and improves the $F_1$ score. Together with the qualitative behavior in Fig.~\ref{fig:dubins}, these results show that the RGB-only controller fails not because the true dynamics are complex, but because the observation space omits the safety-relevant state variable needed for safe control.

\subsection{Thermal Experiments WM and Latent Safety Filters}
\label{app:dreamer-rssm}

\subsubsection{Thermal Unicycle}

The Thermal Unicycle experiments use the same DreamerV3-style RSSM backbone as the \wax~experiments. The model is trained offline from paired observation-action trajectories and learns a recurrent latent state from RGB-only or RGB+IR observations. The resulting latent states are used by the safety margin classifier and latent HJ safety filter described in Sec.~\ref{app:hj-filter}.

\subsubsection{\wax}

For the \wax~experiments, we use a DreamerV3-style recurrent state-space model (RSSM) trained offline from collected robot trajectories. The model encodes high-dimensional observations into an embedding, updates a recurrent latent state using the previous action and observation embedding, and trains a decoder to reconstruct observations from the resulting latent features. The latent feature is formed by concatenating the deterministic recurrent state with the stochastic latent state, and this feature is used both for prediction and for training the safety margin classifier.

We train three variants of the \wax~WM. The RGB-only model encodes and reconstructs only RGB observations. The multimodal model encodes and reconstructs both RGB and infrared observations, using the IR stream to provide direct information about the wax temperature. Finally, the privileged-supervision model receives only RGB observations as input but is trained to reconstruct both RGB and IR observations. This last model does not add sensing at deployment time; instead, the IR reconstruction objective shapes the latent representation during training so that the RGB-conditioned latent dynamics better capture safety-relevant thermal structure.

All \wax~WMs use Gaussian stochastic latents with a 512-dimensional deterministic state and a 32-dimensional stochastic state. RGB and IR images are resized to $224 \times 224$, and actions are represented as 7-dimensional robot commands consisting of delta translation, axis-angle rotation, and gripper action. Although the physical task constrains the robot primarily to vertical motion, we record and normalize the full action vector to lie in $[-1, 1]$. Data is collected at 15Hz and downsampled to 3Hz to match the slow thermal dynamics of wax melting. The learned latent state is then used to train the safety margin classifier and synthesize the latent safety filter.

\begin{table}[ht]
    \centering
    \small{
    \renewcommand{\arraystretch}{1.1}{
    \resizebox{0.9\linewidth}{!}{
    \begin{tabular}{lcc}
        \toprule
        \textbf{\textsc{Hyperparameter}} 
        & \textbf{\textsc{Thermal Unicycle}} 
        & \textbf{\textsc{\wax}} \\
        \midrule
        \textsc{RGB Image Dimension} & [128, 128, 3] & [224, 224, 3] \\
        \textsc{IR Image Dimension} & [128, 128, 1] & [224, 224, 1] \\
        \textsc{Action Dimension} & 2 & 7 \\
        \textsc{Stochastic Latent} & Gaussian & Gaussian \\
        \textsc{Latent Dim (Deterministic)} & 512 & 512 \\
        \textsc{Latent Dim (Stochastic)} & 32 & 32 \\
        \textsc{Activation Function} & SiLU & SiLU \\
        \textsc{Encoder CNN Depth} & 32 & 32 \\
        \textsc{Encoder MLP Layers} & 5 & 5 \\
        \textsc{Failure Projector Layers} & 2 & 2 \\
        \textsc{Batch Size} & 16 & 16 \\
        \textsc{Batch Length} & 64 & 64 \\
        \textsc{Optimizer} & Adam & Adam \\
        \textsc{Learning Rate} & 1e-4 & 1e-4 \\
        \textsc{$\marginfunc(\latent)$ Learning Rate} & 5e-4 & 5e-4 \\
        \textsc{Iterations} & 15000 & 20000 \\
        \bottomrule
    \end{tabular}
    }
    }}
    \vspace{0.5em}
    \caption{Dreamer RSSM hyperparameters for the Thermal Unicycle and \wax~experiments.}
    \label{tab:rssm_hyperparams}
    \vspace{-0.2in}
\end{table}

\subsubsection{Failure Classifier}
The safety classifier $\marginfunc(\latent)$ outputs a scalar value and is trained via a hinge loss parameterized by a  constant $\delta >0$:
\begin{equation}
\begin{aligned}
     \mathcal{L}(\ellparam) = \frac{1}{N_\text{safe}}\sum_{\obs^+ \in \mathcal{D}_{\text{safe}}} \text{ReLU}\big(\delta - \marginfunc(\enc(\obs^+))\big) +\frac{1}{N_\text{fail}}\sum_{\obs^- \in \mathcal{D}_\text{fail}} \text{ReLU}\big(\delta + \marginfunc(\enc(\obs^-))\big), 
     \label{eq:failure-classifier-loss}
\end{aligned}
\end{equation}
where $\mathcal{D} = \mathcal{D}_\text{safe} \cup \mathcal{D}_\text{fail}$ and $o^+$ and $o^-$ refer to observations assigned safe and unsafe labels, respectively. We backpropagate this loss through the entire WM which, in theory, ensures that the latent representation can also identify safety constraints. 
This objective imbues the failure classifier with semantic meaning such that $\marginfunc(\latent_t) < 0 \iff l_t = \texttt{unsafe}$. 

\subsubsection{Latent Safety Filter Synthesis}
\label{app:hj-filter}
All Thermal Unicycle and \wax~safety filters are obtained by approximately solving for the fixed point of a latent Hamilton-Jacobi Bellman equation via an off-policy DDPG algorithm \cite{lillicrap2019continuouscontroldeepreinforcement}. We parameterize the state-action value function and co-optimized safety fallback policy as MLPs trained via rollouts in the latent space of the WM.

Because we do not have access to a simulator, we reset the environment by uniformly sampling random $N=5$ length trajectory segments from the offline dataset to initialize the latent state before performing open-loop imagination rollouts for up to 16 timesteps. This reset procedure and short-horizon rollout reduce instability in the WM imagination. Following \cite{nakamura2025generalizing}, we take $\tanh(\lambda \marginfunc(\latent))$ with $\lambda = 0.1$ as the margin function for reachability learning to smooth the loss landscape.

We train each policy for 600k iterations and save checkpoints every 40k iterations. Because RL training has minor instabilities, we report results from the best-performing checkpoint for each model rather than necessarily using the final checkpoint. We list the DDPG hyperparameters in Table~\ref{tab:DDPG_hyperparams}.

\begin{table}[h]
    \centering
    \resizebox{0.5\linewidth}{!}{
    \renewcommand{\arraystretch}{1.1}
    \begin{tabular}{lc}
        \toprule
        \textsc{\textbf{Hyperparameter}} & \textsc{\textbf{Value}} \\
        \midrule
        \textsc{Actor Architecture} & [512, 512, 512] \\
        \textsc{Critic Architecture} & [512, 512, 512] \\
        \textsc{Normalization} & LayerNorm \\
        \textsc{Activation} & ReLU \\
        \textsc{Discount Factor} $\gamma$ & 0.9999 \\
        \textsc{Learning Rate (Critic)} & 1e-3 \\
        \textsc{Learning Rate (Actor)} & 1e-4 \\
        \textsc{Optimizer} & AdamW \\
        \textsc{Number of Iterations} & 600000 \\
        \textsc{Replay Buffer Size} & 40000 \\
        \textsc{Batch Size} & 512 \\
        \textsc{Max Imagination Steps} & 16 \\
        \bottomrule
    \end{tabular}
    }
    \vspace{0.5em}
    \caption{DDPG hyperparameters for latent HJ safety filter synthesis.}
    \label{tab:DDPG_hyperparams}
    \vspace{-0.2in}
\end{table}

\subsection{Tactile WM and Model-Predictive Shielding}
\label{app:tactile-mps}

\subsubsection{\rice: ViT-based WM}
\label{subsub:rice}

We design a ViT-based WM following DINO-WM \cite{zhou2024dino}. RGB observations are encoded with DINOv3-S \cite{simeoni2025dinov3}, producing dense visual token embeddings of size $196 \times 384$. GelSight tactile observations are encoded with AnyTouch 2 \cite{feng2026anytouch}, producing tactile token embeddings of size $196 \times 512$. These embeddings define the latent representations for the RGB and tactile modalities.

To model dynamics, we condition on a history of $H=3$ time steps consisting of DINO embeddings, end-effector states, and actions. At each timestep, we concatenate the end-effector state $(8\text{D})$ and an encoded action embedding $(10\text{D})$ to each DINO token, yielding an input latent sequence of size $3 \times 196 \times 402$. To incorporate tactile context, we take the most recent AnyTouch embedding, mean-pool it over the token dimension, and project it to a 128-dimensional conditioning vector. This tactile vector modulates the ViT dynamics model through AdaLN conditioning. The model predicts a $3 \times 196 \times 512$ latent state, which is then projected to future DINO token embeddings and end-effector states. We train the WM with one-step teacher forcing. For RGB-only WMs, we keep the same architecture and replace the tactile input with a zero-valued $196 \times 512$ tensor.

\subsubsection{Failure Classifier}

For failure classification, we mean-pool the predicted $196 \times 512$ latent state over the patch dimension and concatenate it with the tactile conditioning vector, forming a 640-dimensional representation. This representation is passed through an MLP to produce a failure logit. The failure classifier is trained with a binary cross-entropy loss on the spill labels. We do not apply temperature scaling or otherwise calibrate the classifier logits directly. Instead, calibration enters through the model-predictive shielding threshold: for each classifier, we tune the intervention threshold $\epsilon$ using the held-out conformal calibration procedure described in Sec.~\ref{app:conformal}. 

We train the failure classifier on WM imagined 16-timestep trajectories: the model observes the first $H=3$ frames and then predicts the remaining 13 timesteps open-loop. During open-loop prediction, the tactile input is held fixed to avoid leakage from future tactile observations. This is a reasonable approximation in our setting because tactile images evolve slower than RGB observations. As a result, the failure classifier is trained on the same type of open-loop rollouts produced by the WM at inference time. 

\subsubsection{Model-Predictive Shielding}

We use a model-predictive shielding controller for our \rice~experiments. We collect data at 15Hz and train the WM described in Sec.~\ref{subsub:rice} using a frame skip of 2. The diffusion policy is trained directly on the 15Hz data and is trained to output action chunks of 16 timesteps. We subsample every other timestep from the action chunk and generate an 8-step rollout in the WM, corresponding to the 16-timestep action chunk from the diffusion policy. We evaluate the worst-case failure probability over these 8 predicted frames and compare it to the threshold $\epsilon$ chosen by the conformal calibration procedure. If the failure probability is less than $\epsilon$, we execute the first 8 actions from the diffusion policy and replan (corresponding to the first half of the action chunk). Otherwise, the MPS controller overrides the nominal action sequence and executes the fallback policy (e.g. zero end-effector motion).

We apply the same MPS procedure to both the RGB and multimodal WMs. The RGB model only receives visual observations, while the multimodal model additionally receives tactile information. To mitigate prediction gaps, we calibrate the shielding threshold using held-out calibration rollouts. This yields a more conservative threshold for the RGB model, whose latent predictions are less informative about whether the opaque bottle contains rice, and a less conservative threshold for the multimodal model. 

We train the WM for 20k iterations, the failure classifier for 1k iterations, and the diffusion policy for 10k iterations. The base WM and diffusion policy are trained using a cosine learning rate schedule with linear warmup. We list the relevant hyperparameters in Tables~\ref{tab:tacwm_hyperparams} and~\ref{tab:mps_hyperparams}.

\begin{table}[t]
    \centering
    \small
    \setlength{\tabcolsep}{4pt}
    \renewcommand{\arraystretch}{1.05}

    \begin{minipage}[t]{0.49\linewidth}
        \vspace{0pt}
        \centering
        \resizebox{\linewidth}{!}{
        \begin{tabular}{lc}
            \toprule
            \textbf{Hyperparameter} & \textbf{Value} \\
            \midrule
            Image size & 256 \\
            DINOv3 patch size & $(14 \times 14, 384)$ \\
            AnyTouch2 patch size & $(14 \times 14, 512)$ \\
            Predictor lr & $1\mathrm{e}{-4}$ \\
            Action encoder lr & $2\mathrm{e}{-4}$ \\
            Action emb. dim & 10 \\
            Proprio. emb. dim & 8 \\
            Batch size & 32 \\
            Batch len. & 4 \\
            EMA & 0.999 \\
            Frame skip & 2 \\
            Warmup iters. & 2000 \\
            Training iters. & 20000 \\
            ViT depth & 6 \\
            ViT heads & 16 \\
            ViT MLP dim & 2048 \\
            Optimizer & AdamW \\
            \bottomrule
        \end{tabular}
        }
        \vspace{0.5em}
        \captionof{table}{\rice~WM hyperparameters.}
        \label{tab:tacwm_hyperparams}
    \end{minipage}
    \hfill
    \begin{minipage}[t]{0.49\linewidth}
        \vspace{0pt}
        \centering
        \resizebox{\linewidth}{!}{
        \begin{tabular}{lc}
            \toprule
            \textbf{Hyperparameter} & \textbf{Value} \\
            \midrule
            State normalization & Yes \\
            Action normalization & Yes \\
            Action chunk & 16 \\
            Image chunk & 2 \\
            Image size & 256 \\
            Batch size & 100 \\
            Training iters. & 10000 \\
            Learning rate & $1\mathrm{e}{-4}$ \\
            LR schedule & Cosine \\
            Optimizer & AdamW \\
            \midrule
            Calibration trajectories & 50 \\
            Calibration quantile $\alpha$ & 0.12 \\
            RGB threshold $\epsilon$ & 0.02 \\
            Multimodal threshold $\epsilon$ & 0.24 \\
            \bottomrule
        \end{tabular}
        }
        \vspace{0.5em}
        \captionof{table}{\rice~diffusion policy hyperparameters.}
        \label{tab:mps_hyperparams}
    \end{minipage}
\end{table}

\end{document}